%% file: CuNAAarxiv.tex
\documentclass[journal]{IEEEtran}

\usepackage{cite}
\usepackage{amsmath,amssymb,amsfonts,amsthm}
\usepackage{algorithmic}
\usepackage{graphicx}
\usepackage{textcomp}
\usepackage{wrapfig}
\usepackage{bm}

\newcommand{\acceleration}[2]{\bm{a}_{#1}}

\newcommand{\aReading}[1]{\widehat{\bm{a}}_{#1}}
\newcommand{\Da}{D_{\alpha}}
\newcommand{\Dw}{D_{\Omega^2}}
\newcommand{\Rdiff}{S_{d}}
\newcommand{\Vicon}{Vicon\textregistered}
\newcommand{\Sec}{Sec. }
\newcommand{\Fig}{Fig. }
\newcommand{\Eqn}{Eqn. }
\newcommand{\Appen}{Appendix }

\begin{document}
\title{Angular Velocity Estimation using Non-coplanar Accelerometer Array}
\author{Michael~Maynard,~\IEEEmembership{Student Member,~IEEE,}
        and~Vishesh~Vikas,~\IEEEmembership{Member,~IEEE}
\thanks{M. Maynard and V. Vikas are with the Agile Robotics Lab (ARL), University of Alabama, Tuscaloosa, AL 35406 USA. %
e-mail: mcmaynard@crimson.ua.edu, vvikas@ua.edu.}
}
\maketitle


%
\input{00-Abstract}
\IEEEpeerreviewmaketitle

%
%
%
%
\input{01-Introduction}
%
\input{02-ProblemDefinition}

\input{03-System}

\input{04-EKF}
\input{05-Conclusion}
\section*{Acknowledgment}
The authors would like to thank Dr. Dario Martelli for help with use of \Vicon~tracking system.

\input{06-Appendices}

\bibliographystyle{IEEEtran}
\bibliography{IEEEabrv,references}
\end{document}

%% file: 00-Abstract.tex
\begin{abstract}
Over the last few decades, Gyro-Free Inertial Measurement Units (GF-IMUs) have been extensively researched to overcome the limitations of gyroscopes. This research presents a Non-coplanar Accelerometer Array (NAA) for estimating angular velocity with non-specific geometric arrangement of four or more triaxial accelerometers with non-coplanarity constraint. The presented proof of non-coplanar spacial arrangement also provides insights into propagation of the sensor noise and construction of the noise covariance matrices. The system noise depends on the singular values of the relative displacement matrix (between the sensors). A dynamical system model with uncorrelated process and measurement noise is proposed where the accelerometer readings are used simultaneously as process and measurement inputs. The angular velocity is estimated using an Extended Kalman Filter (EKF) that discretizes and linearizes the continuous-discrete time dynamical system. The simulations are performed on a Cube-NAA (Cu-NAA) comprising four accelerometers placed at different vertices of a cube.%
They analyze the estimation error for static and dynamic movement as the distance between the accelerometers (four accelerometers in cube-orientation) is varied. Here, the system noise is observed to decrease inversely with the length of the cube edge as the arrangement is kept identical. Consequently, the simulation results indicate asymptotic decrease in the standard error of estimation with edge length. The experiments are conducted on a Cu-NAA with five reflective optical markers. The reflective markers are visually tracked using \Vicon~to construct the ground truth angular velocity. This unique experimental setup, apart from providing three degrees of rotational freedom of movement, also allows for three degrees of spacial translation (linear acceleration of the Cu-NAA in space). The simulation and experimental results indicate better performance of the proposed EKF as compared to one with correlated process and measurement noises.
\end{abstract}

\begin{IEEEkeywords}
GF-IMU, gyro-free IMU, angular velocity estimation, sensor fusion, accelerometer array.
\end{IEEEkeywords}

%% file: 01-Introduction.tex
\section{Introduction}

\IEEEPARstart{T}{raditional} navigation systems utilize measurements from external electromagnetic radiation (light or radio waves), or earth's magnetic field (magnetic compass). On the contrary, inertial navigation systems tend to operate without reliance on information about external fields or radiation. Unfortunately, navigation relying upon external radiation is susceptible to changes in conditions and to accidental or intentional interference which may degrade or destroy its effectiveness (jamming). From this perspective, inertial navigation systems are independent of weather, visibility and terrain conditions as they do not require external signals or emit radiation (impossible to detect). More so, once initialized, they can function automatically without human intervention.

However, the limitations of such systems relate to the inertial sensors themselves. 
Microelectromechanical Sensors (MEMS) inertial sensors are widely popular due to economic and design reasons. MEMS gyroscopes directly measure angular velocity and differ from traditional gyroscopes by utilizing principle of vibration of mass on a turntable to eliminate need for bearing-like mechanical parts. Contrastingly, MEMS accelerometers are passive sensors that measure acceleration by utilizing the beam bending principles. These design differences provide MEMS accelerometers with advantages including lower power consumption, cost, weight, broader dynamic range and shorter reaction time \cite{chen_gyroscope_1994}. %
Unlike MEMS accelerometers, the MEMS gyroscopes are prone to bias instabilities over time due to multiple factors including temperature. This can be compensated using references from magnetometer or accelerometer (zero angular velocity update) of the Inertial Measurement Unit (IMU). Consequently, the underlying motivation of designing Gyroscope-Free Inertial Measurement Units (GF-IMUs) is to translate the advantages and precision of MEMS accelerometers to construct more precise angular velocity estimators. The estimation of angular velocity is made possible from the fact that the acceleration of a point on a rigid body is proportional to the square of the angular velocity vector and angular acceleration of the body\cite{tan_design_2001,tan_feasibility_2000,barbour_inertial_2001}. The research in the field of design and analysis of GF-IMUs has been extensive over the last few decades starting from six linear accelerometers \cite{chen_gyroscope_1994, park_scheme_2005} placed in restrictive geometric orientations e.g. cube configuration, for estimating angular velocity. Thereafter, optimal geometric designs were explored \cite{hanson_optimal_2005,zappa_number_2000}, and Extended Kalman Filter (EKF), Unscented Kalman Filter (UKF) techniques were applied to estimate the angular velocity \cite{cardou_angular_2008,schopp_design_2010,schopp_observing_2014,edwan_constrained_2011,lu_state_2011}.  The reader may refer to \cite{skog_inertial_2016} for more details on inertial sensor arrays. However, the dynamic models have correlated process and measurement noise. Furthermore, the experimental validation of these algorithms are performed using a setup where the accelerometer array only rotates in space and does not linearly accelerate (rotation tables). 

\textit{Contributions}: We present a distributed sensor network of Non-coplanar Accelerometer Array (NAA) for estimating angular velocity with (i) Non-specific geometric arrangement of accelerometers: We provide proof for non-specific geometric arrangement of four or more triaxial accelerometers with the constraint of non-coplanar placement. The analysis reveals that the propagation of the sensor noise depends on the relative displacement matrix composed of concatenated displacement vectors between sensors. More specifically, its condition number and product of the singular values. Ideally, a condition number of one and high product of singular values is desired to minimize propagation of sensor noise.  (ii) EKF with uncorrelated process and measurement noise: For estimation, the proposed dynamical system model has uncorrelated process and measurement noise where the accelerometer readings are used both as measurement and process input. Consequently, the angular velocity is estimated using an EKF that discretizes and linearizes the continuous-discrete time system. The simulation analyzes the estimation error for static and dynamic movement as the distance between the accelerometers (four accelerometers in cube-geometry) is varied. The results  indicate  asymptotic decrease in the  standard  error  of  estimation  with  edge  length. (iii) Experimental setup permitting linear acceleration: The unique experimental setup permits linear acceleration of the multi-accelerometer sensor in space where the ground truth is assumed to be obtained from the optical \Vicon tracking system. 

The paper is structured as follows: \Sec \ref{Sec:ProbDef} defines the problem including nomenclature for sensor array. This section also discusses the proof for need of minimum four non-coplanar accelerometer for calculating the angular velocity. Next, the continuous-discrete dynamical system model and the discrete EKF are derived in \Sec \ref{Sec:Model}. Finally, the \Sec \ref{Sec:Results} discusses the simulation and experimental setup, and their results.

%% file: 02-ProblemDefinition.tex
\section{Problem Definition} %
\label{Sec:ProbDef}

\begin{figure}
\begin{center}
\includegraphics[trim={5pt 5pt 5pt 10pt},clip,width=0.5\columnwidth]{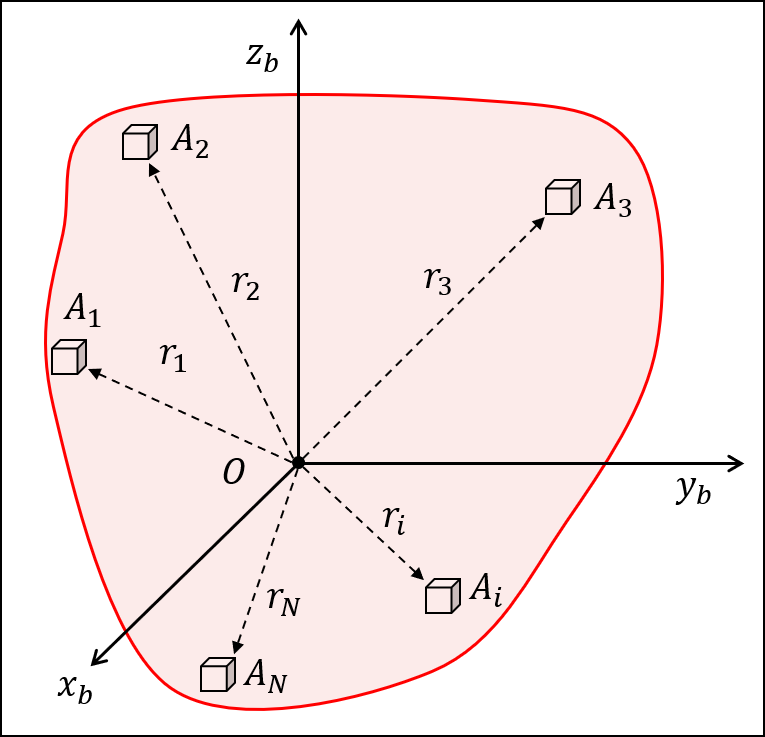}
\caption{The accelerometers $A_i$ are located on a rigid body (shaded red) at a distance $\bm{r}_i$ from the origin $O$. The $\{x_b,y_b,z_b\}$ are orthonormal basis of the body coordinate system.}
\label{Fig:SoGyro}
\end{center}
\end{figure}

Let body reference frame $\{b\}$ with origin $O$ and orthonormal basis vectors $\{x_b,y_b,z_b\}$, rotate with angular velocity and acceleration of $\bm{\omega}=\left[\omega_1,\omega_2,\omega_3\right]^T$, $\bm{\alpha}=\left[\alpha_1,\alpha_2,\alpha_3\right]^T$ respectively. Assume, $N$ accelerometers are placed at $\bm{r}_{i}, i=1,\cdots,N$ on the rigid body, \Fig \ref{Fig:SoGyro}. It is desired to (i) identify the minimum number of accelerometers required to estimate the angular velocity and (ii) estimate the angular velocity given the relative distance between the accelerometers.

Theoretically, the matrix representation of acceleration $\acceleration{i}{}$ measured by accelerometer $A_i\ \forall i\in[1,N]$ is
\begin{align}
\acceleration{i}{} &= \acceleration{O}{} + \bm{\alpha} \times \bm{r}_i + \bm{\omega}\times \left(\bm{\omega} \times \bm{r}_i\right) \nonumber\\
&= \acceleration{O}{} + D\left(\bm{r}_i\right) \bm{y}  \label{Eqn:SensorEqn}\\
D\left(\bm{r}\right) &= 
\left[\begin{array}{c@{\hspace{2pt}}c@{\hspace{2pt}}c@{\hspace{2pt}}c@{\hspace{2pt}}c@{\hspace{2pt}}c@{\hspace{2pt}}c@{\hspace{2pt}}c@{\hspace{2pt}}c@{\hspace{2pt}}}
0 & -r_1 & -r_1  & 0 & r_3 & r_2 & 0 & r_3 & -r_2\\
-r_2 & 0&-r_2 & r_3 & 0  & r_1 & -r_3 & 0 & r_1\\
-r_3 & -r_3 & 0& r_2 & r_1  & 0 & r_2 & -r_1 & 0
\end{array}
\right]
\label{Eqn:theDfunc}\\
\bm{y} &= \left[\omega_1^2, \omega_2^2, \omega_3^2, \omega_2\omega_3, \omega_3\omega_1, \omega_1\omega_2, \alpha_1, \alpha_2, \alpha_3\right]^T \label{Eqn:thesmallx}
\end{align}
where $\bm{r}=[r_1,r_2,r_3]^T$, and $\bm{a}_O$ is the acceleration of point $O$.

\noindent \textit{Sensor Array.} Let the accelerometer $A_i$ have noise $\bm{e}_i$ with covariance matrix $Q_i$. The sensor measurement $\aReading{i}$ is
\begin{align}
    \aReading{i} = \acceleration{i}{} + \bm{e}_i, \qquad \mathbb{E}\left[\bm{e}_i \bm{e}_i^T\right]=Q_i
\end{align}
In context of the sensor array, the measurement $\mathbf{\widehat{a}}$, acceleration $\mathbf{a}$ and noise $\mathbf{e}$ column vectors, and the noise convariance matrix $Q$ are defined as
\begin{align}
    \mathbf{\widehat{a}} &= \left[\aReading{1}^T,\aReading{2}^T,\cdots,\aReading{N}^T\right]^T, \quad
    \mathbf{{a}} = \left[\acceleration{1}{}^T,\acceleration{2}{}^T,\cdots,\acceleration{N}{}^T\right]^T \nonumber\\
    \mathbf{e} &= \left[\bm{e}_1, \bm{e}_2,\cdots,\bm{e}_N\right]^T, \qquad \mathbf{\widehat{a}, a, e} \in \mathbb{R}^{3N\times1}\\ 
    Q &= \mathbb{E}[\mathbf{e}\mathbf{e}^T] = \mathrm{diag}\left(Q_1,Q_2,\cdots,Q_N\right), \quad Q \in \mathbb{R}^{3N\times 3N} \nonumber
\end{align}
To facilitate compact and elegant representation of relative accelerations, we define matrices $E \in \mathbb{R}^{3(N-1)\times N}, G \in \mathbb{R}^{3(N-1)\times 9}$
\begin{align}
    E &= \begin{bmatrix}
    I &-I & 0 & \cdots &0 & 0\\
    0 & I & -I & \cdots &0 & 0\\
    \vdots & \ddots & \ddots & \vdots\\
    0 & \cdots & \cdots & \cdots & I & -I
     \end{bmatrix} \label{Eqn:EMat}\\
\mathrm{s.t.}\quad     E\mathbf{\widehat{a}} &=
    \begin{bmatrix}
    \aReading{1}-\aReading{2}\\
    \aReading{2}-\aReading{3}\\
    \vdots\\
    \aReading{N-1}-\aReading{N}
    \end{bmatrix} \nonumber\\
    G &= \begin{bmatrix}
    D(\bm{r}_1-\bm{r}_2)\\
    D(\bm{r}_2-\bm{r}_3)\\
    \vdots\\
    D\left(\bm{r}_{(N-1)}-\bm{r}_N\right)
    \end{bmatrix} \label{Eqn:GMat}
\end{align}
Consequently,
\begin{align}
    E\mathbf{\widehat{a}} &= E(\mathbf{a-e}) \quad
    \Rightarrow \quad E\mathbf{\widehat{a}} = G\bm{y} - E\bm{e} \label{Eqn:AccelDiff}
\end{align}

\noindent \textit{Proposition:} At a given time moment, minimum of four non-coplanarly placed accelerometers are needed to calculate vector $\bm{y}$.
\\\begin{proof}
Given $N$ accelerometers, the relative difference in acceleration can be written as
\begin{align*}
    \quad E\mathbf{\widehat{a}} = G\bm{y} - E\bm{e}
\end{align*}
The least squares solution $\bm{y}^*$ exists only if $G$ is full ranked. The invertability is examined by performing elementary row operations on matrix $D(\bm{r})$, hence, $G$. The matrix $\widetilde{D}\left(\mathbf{r}\right)$ is obtained by performing the following column operations on columns of $D(\bm{r})$ : %
$\widetilde{C}_{1}\leftarrow \left(C_1-C_2-C_3\right)/2$, %
$\widetilde{C}_{2}\leftarrow \left(C_4-C_9\right)/2$, %
$\widetilde{C}_{3}\leftarrow \left(C_6+C_8\right)/2$, %
$\widetilde{C}_{4}\leftarrow \left(C_4+C_9\right)/2$, %
$\widetilde{C}_{5}\leftarrow -\left(C_1-C_2+C_3\right)/2$, %
$\widetilde{C}_{6}\leftarrow \left(C_5-C_7\right)/2$, %
$\widetilde{C}_{7}\leftarrow \left(C_6-C_8\right)/2$, %
$\widetilde{C}_{8}\leftarrow \left(C_5+C_7\right)/2$, %
$\widetilde{C}_{9}\leftarrow -\left(C_1+C_2-C_3\right)/2$ %
where $\widetilde{C}_i,C_j \forall i,j$ correspond to columns of matrices $\widetilde{D}(\bm{r}),D(\bm{r})$ respectively.
\begin{align*}
\widetilde{D}\left(\mathbf{r}\right) = %
\begin{bmatrix}
\bm{r}^T & 0 & 0\\
0 & \bm{r}^T & 0\\
0 & 0 & \bm{r}^T
\end{bmatrix}
\end{align*}
Thereafter, performing row interchanging operation on the cumulative $G$ to obtain $\widetilde{G}$: $\widetilde{R}_i\leftarrow R_{3i},  R_{3(N-1)+i}\leftarrow R_{3i+1}, \widetilde{R}_{6(N-1)+i}\leftarrow R_{3i+2}, \forall i=1,2,\cdots,(N-1)$ where $\widetilde{R}_i,R_j$ correspond to rows of matrices $\widetilde{G}, G$
\begin{align}
\widetilde{G} = 
\begin{bmatrix}
\Rdiff & 0 & 0\\
0 & \Rdiff &0\\
0 & 0 & \Rdiff
\end{bmatrix}
\label{Eqn:GTilde}
\end{align}
where the relative displacement matrix $\Rdiff \in \mathbb{R}^{(N-1)\times 3}$
\begin{align}
\Rdiff = \left[ (\bm{r}_1-\bm{r}_2), (\bm{r}_2-\bm{r}_3), \cdots, (\bm{r}_{N}-\bm{r}_{N-1})
\right]^T
\label{Eqn:Rdiff}
\end{align}
Concisely, let full-ranked matrices $A\in \mathbb{R}^{3(N-1)\times 3(N-1)}, B\in \mathbb{R}^{9 \times 9}$ correspond to the previously discussed elementary row and column operations
\begin{align}
    G = A\widetilde{G}B \label{Eqn:Gdecomposition}
\end{align}
The elementary column and row operations imply $\mathrm{rank}(G) = \mathrm{rank}(\widehat{G}) = 3 \times \mathrm{rank}(\mathbf{r}_{d})$ where $\mathrm{rank}(\mathbf{r}_{d})\leq 3$. Hence, $G$ is full-ranked only when $\mathbf{r}_{d}$ is full-ranked i.e. rank is 3. This condition is fulfilled in the cases when at least three of the displacement vector combinations,  $(\bm{r}_1-\bm{r}_2), (\bm{r}_2-\bm{r}_3), \cdots, \left(\bm{r}_{N}-\bm{r}_{(N-1)}\right)$ are linearly independent. In other words, the accelerometers are placed in a non-coplanar orientation.
Hence, it can be concluded that a minimum number of $N=4$ non-coplanarly placed accelerometers are required to obtain vector $\bm{y}$.
\end{proof}

The $E,G$ matrices in \Eqn {\ref{Eqn:EMat}}, {\ref{Eqn:GMat}} consider set of $(N-1)$ relative difference between consecutive accelerometers. However, one may exhaustively use $\displaystyle N_{all}=\frac{N!}{2(N-2)!}$ combinations between accelerometers. These additional elements are linear combinations of the previous set, hence, such calculations will not alter the algorithmic results.

\textit{Sensor noise propagation.} The least squares solution $\bm{y}^*$ to Eqn \ref{Eqn:AccelDiff} is
\begin{equation}
\bm{y}^* = G^{+}E(\mathrm{a-e}) \label{Eqn:ystar}
\end{equation}
The behavior of the transformed noise $G^{+}E \mathbf{e}$ is dictated by the relative displacement matrix $\Rdiff$. As matrices $A,B,E$ are constant, this is evident
\begin{align*}
    G^{+} &= B^{-1}\widetilde{G}^{+}A^{-1}\\
    \widetilde{G}^{+} &= \begin{bmatrix}
    {\Rdiff}^{+} & 0 & 0\\
    0 & {\Rdiff}^{+} & 0\\
    0 & 0 & {\Rdiff}^{+}
    \end{bmatrix}
\end{align*}
Consequently, it can be observed that the singular values of $\Rdiff$, particularly the condition number $\mathrm{cond}(\Rdiff)$ and product of the singular values $\sqrt{\det(\Rdiff^T\Rdiff)}$ determine the behavior of the transformed noise $G^{+}E\bm{e}$. Geometrically, the most desirable orientation would be  $\mathrm{cond}(\Rdiff)\rightarrow 1$ with large relative distance, i.e., high $\det(\Rdiff^T\Rdiff)$.

%% file: 03-System.tex
\section{Dynamical System Model and Extended Kalman Filter} \label{Sec:Model}
We propose a continuous-discrete time dynamical model with uncorrelated process and measurement noise with states  $\bm{x}=[\omega_1,\omega_2,\omega_3]^T$. 
\begin{align}
    \dot{\bm{x}} &= M\mathbf{a}-Lh(\bm{x}) +M\mathbf{e} \label{Eqn:Sysprocess}\\
    \bm{z} &= h(\bm{x}) + \Dw\mathbf{e} \label{Eqn:Sysmeasurement}
\end{align}
where $\bm{z},h(\bm{x})\in\mathbb{R}^{6 \times 1}$ and 
\begin{align}
    G^+E &=
    \begin{bmatrix}
    \Dw\\
    \Da
    \end{bmatrix}, \quad \Dw\in \mathbb{R}^{6\times N}, \Da\in \mathbb{R}^{3\times N} \nonumber\\
    L &= -\left(\Da Q \Dw^T\right)\left(\Dw Q \Dw^T\right)^{-1}\\
    M & = \Da + L\Dw \nonumber\\
    h(\bm{x}) &=[x_1^2,x_2^2,x_3^2,x_1 x_2,x_2 x_3,x_3 x_1]^T \nonumber
\end{align}
\textit{Derivation.} The vector $\bm{y}$ is a vector containing angular acceleration $\alpha_i$ and second order angular velocity terms $\omega_i\omega_j$ and $\forall i,j=1,2,3$. For the defined states, $\displaystyle \bm{y} = [h(x)^T,\bm{\alpha}^T]^T$. The least squares solution for $\bm{y}$ is
\begin{align*}
    \bm{y} &= G^{+}E(\mathbf{a-e}) \Rightarrow 
    \begin{bmatrix}
    h(\bm{x})\\
    \bm{\alpha}
    \end{bmatrix} = 
    \begin{bmatrix}
    \Dw (\mathbf{a-e})\\
    \Da(\mathbf{a-e})
    \end{bmatrix}
\end{align*}
Consequently, we can construct the state dynamics and measurement as
\begin{align*}
    \dot{\bm{x}} &= \underbrace{\Da(\mathbf{a-e})}_{=\bm{\alpha}}\\
    \underbrace{\bm{z}}_{=\Dw\mathbf{a}}&=  h(\bm{x}) +\Dw\mathbf{e}
\end{align*}
However, the state and measurement noise are not uncorrelated, i.e., $\displaystyle \mathbb{E}\left[-\Da\mathbf{e}(\Dw \mathbf{e})^T \right] = -\Da Q \Dw^T \neq 0$. The uncorrelated noise $\widetilde{\mathbf{e}}$ is 
\begin{align*}
    \widetilde{\mathbf{e}} &= -\Da \mathbf{e} -L \Dw \mathbf{e}, \quad 
    \mathrm{s.t.} \quad \mathbb{E}\left[\widetilde{\mathbf{e}} (\Dw \mathbf{e})^T\right] = 0\\
    \Rightarrow L &= -\left(\Da Q \Dw^T\right)\left(\Dw Q \Dw^T\right)^{-1}
\end{align*}
and we define $M=\Da + L \Dw$ such that $\widetilde{\mathbf{e}} = M\mathbf{e}$
\begin{align*}
    -\Da \mathbf{e} &= \widetilde{\mathbf{e}} +L\Dw \mathbf{e}\\
    &= \widetilde{\mathbf{e}} + L\left(\Dw \mathbf{a} - h(\bm{x})\right)
\end{align*}
Hence, for uncorrelated state and measurement noise, $\displaystyle \mathbb{E}\left[M\mathbf{e}(\Dw \mathbf{e})^T\right]=0$
\begin{align*}
    \dot{\bm{x}} &= M\mathbf{a}-Lh(\bm{x}) +M\mathbf{e}\\
    \bm{z} &= h(\bm{x}) + \Dw\mathbf{e} 
\end{align*}
\hfill $\blacksquare$

The continuous-discrete time system is discretized and then linearized to construct EKF. We use the \cite{welch_introduction_1995} notation to enable ease of understanding
\begin{align}
    \bm{x}_k &= f(\bm{x}_{k-1}) + \bm{w}_{k-1} \label{Eqn:DiscreteEKFProcess}\\
     &= \left(\bm{x}_{k-1} - L h(\bm{x}_{k-1})T + M \mathbf{a}T\right) + T\widetilde{\bm{e}} \nonumber\\
    \bm{z}_k &= h(\bm{x}_k) + \bm{v}_k \label{Eqn:DiscreteEKFMeasurement}\\
     &= h(\bm{x}_k) + \Dw\mathbf{e} \nonumber
\end{align}
where $T$ is the sample time and the Jacobians and the noise covariance matrices are
\begin{align}
    H(\bm{x}) &= \frac{\partial h(\bm{x})}{\partial \bm{x}} = 
    \begin{bmatrix}
    2x_1 & 0 & 0 & x_2 & 0 & x_1\\
    0 & 2x_2 & 0 & x_1 & x_3 & 0\\
    0 & 0 & 2x_3 & 0 & x_2 & x_1
    \end{bmatrix}^T\\
    F(\bm{x}) &= \frac{\partial f(\bm{x})}{\partial \bm{x}} = I - LH(\bm{x})T
\end{align}
The EKF is implemented in a recursive fashion assuming $P_0, x_0$ at $t=0$
\begin{enumerate}
    \item \textbf{Time update}
    \begin{align*}
        \bm{x}_k^- &= \left(\bm{x}_{k-1} - L h(\bm{x}_{k-1})T + M \mathbf{a}T\right)\\
        P_k^- &= F(\bm{x}_{k-1})P_{k-1}F(\bm{x}_{k-1})^T + T^2MQM^T
    \end{align*}
    \item \textbf{Measurement update}
    \begin{align*}
        K_k &= P_k^{-1} H_k^T(H_kP_k^{-1}H_k^T + R_k)^{-1}\\
        \bm{x}_k &= \bm{x}_k^{-} + K_k \left(\Dw \mathbf{a} - h(\bm{x}_k^-)\right)\\
        P_k &= (I-K_k H_k)P_k^-
    \end{align*}
\end{enumerate}

\begin{figure}
\begin{center}
\includegraphics[trim={5pt 5pt 5pt 10pt},clip,width=0.5\columnwidth]{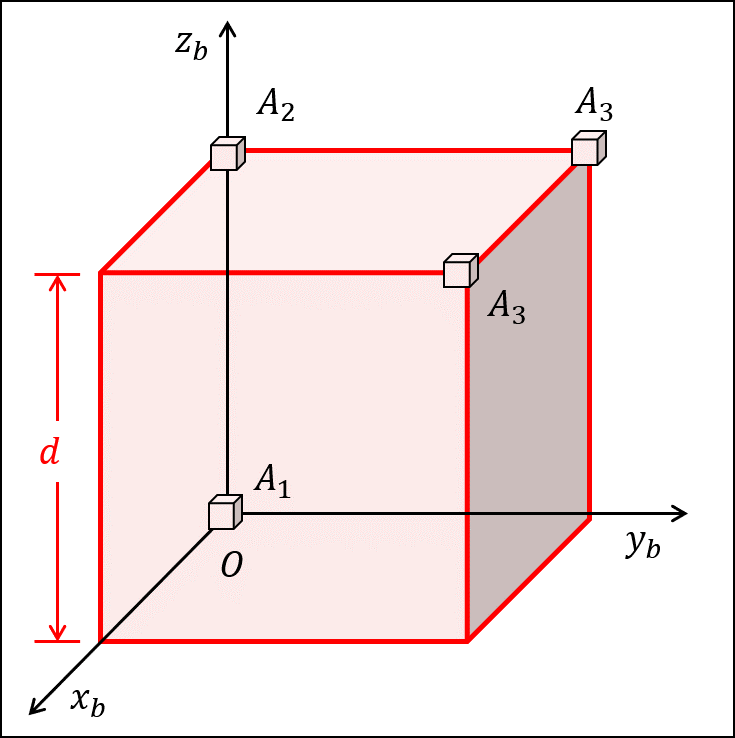}
\caption{The simulation considers Cu-NAA where four accelerometers $A_1,A_2,A_3,A_4$ are placed at the vertices of a cube with edge length $d$. For this arrangement, the relative difference matrix $\Rdiff$ has $\mathrm{cond}(\Rdiff)=1, \sqrt{\mathrm{det}(\Rdiff \Rdiff^T)}= d$.}
\label{Fig:simSoGyro}
\end{center}
\end{figure}

%% file: 04-EKF.tex
\section{Simluation and Experimental Results} \label{Sec:Results}
\subsection{Simulation}
We simulate a Cu-NAA where the accelerometers are placed at the vertices of cube with edge length $d$, \Fig \ref{Fig:simSoGyro}, and 
\begin{align*}
    S_d = d\begin{bmatrix}
    0 & 0 & 1\\
    0 & 1 & 0\\
    1 & 0 & 0
    \end{bmatrix}
\end{align*}
such that $\mathrm{cond}(S_d)=1$ and $\sqrt{\mathrm{det}(S_dS_d^T)}=d$. The simulation is performed in MATLAB\textregistered~ with accelerometer noise of $0.02m/\sec^2$ and sampling frequency of $100~Hz$. We examine the performance of the estimator when the Cu-NAA is dynamically rotated and kept stationary, and as the edge length $d$ is varied. Dynamic rotation is simulated with roll and yaw varying sinusoidally at frequency, amplitude and phase of of $0.5Hz,~ 0.75Hz,~ 10\deg/\sec,~20\deg/\sec,~25\deg$ and $40\deg$ respectively. For the case of $d=10cm$, the EFK estimate has standard error of $1.14, 1.05, 0.97 \deg/\sec$ in $x,y,z$ directions, \Fig \ref{Fig:simSoGyro}. While the standard error when for the stationary case $\omega=[0,0,0]^T$ is {$2.85, 2.66, 2.25 \deg/\sec$ in $x,y,z$ directions with mean of $2.59\deg/\sec$}. In comparison, the EKF estimator with correlated process-measurement noise ($L=\bm{0}_{3\times 6}$ in \Eqn {\ref{Eqn:Sysprocess}}) has standard error of (i) dynamic case: $1.20, 1.08, 1.01 \deg/\sec$, and (ii) static case:  $-2.28, 1.67, 2.12\deg/\sec$.
\begin{figure}[h]
\begin{center}
\includegraphics[trim={150pt 25pt 150pt 50pt},clip,width=\columnwidth]{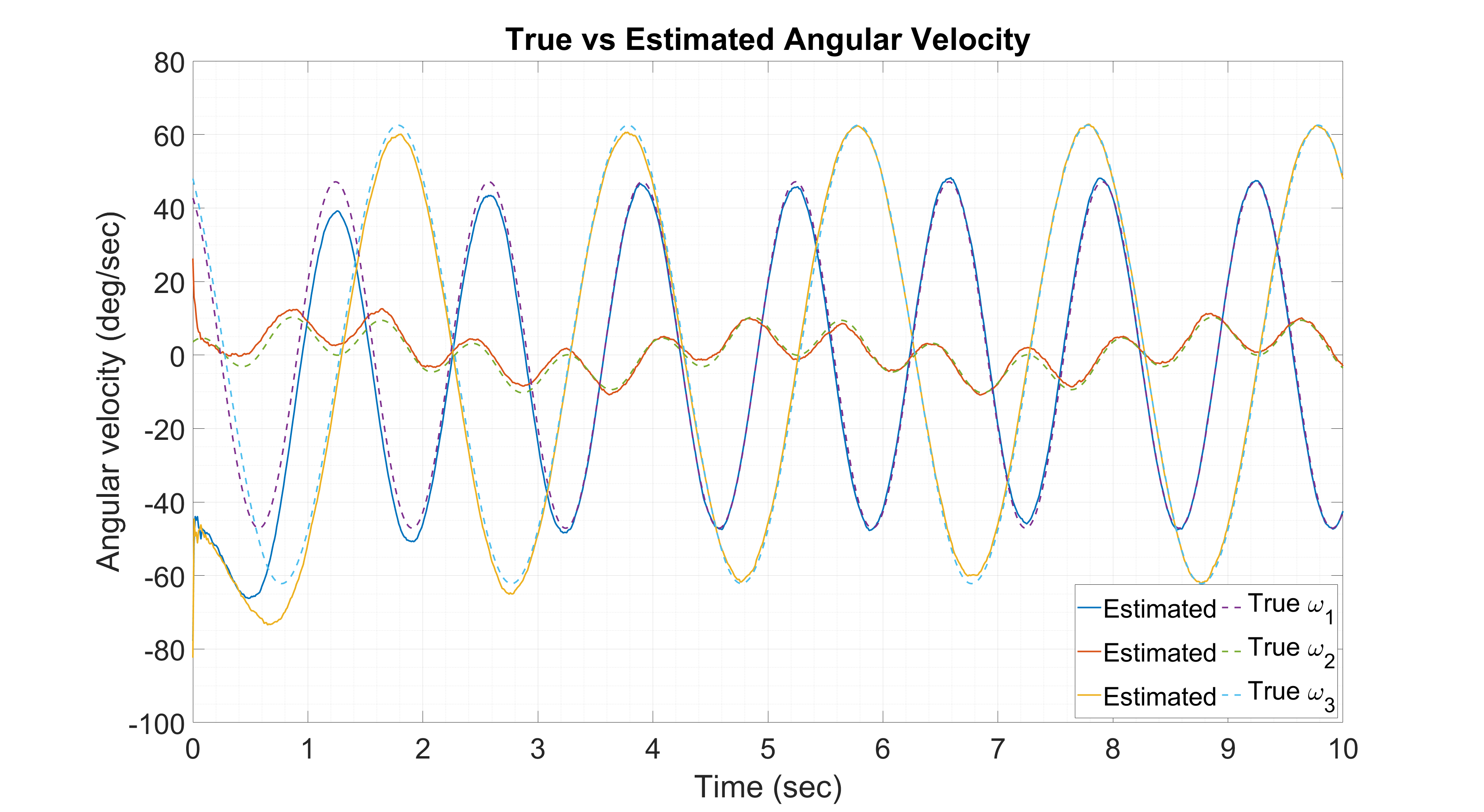}
\caption{Dynamic rotation of Cu-NAA: The estimated angular velocity for $d=10 cm$ (solid lines) track the true angular velocities (dotted lines) with mean standard error of $1.09 \deg/\sec$. {The accelerometer are sampled at $100 Hz$ with Gaussian noise standard deviation of $0.02 m/\sec^2$.}} 
\label{Fig:SoGyroDynamic}
\end{center}
\end{figure}

\begin{figure}
{\scriptsize \textbf{(a)}}
\begin{center}
\includegraphics[trim={30pt 15pt 90pt 25pt},clip,width=\columnwidth]{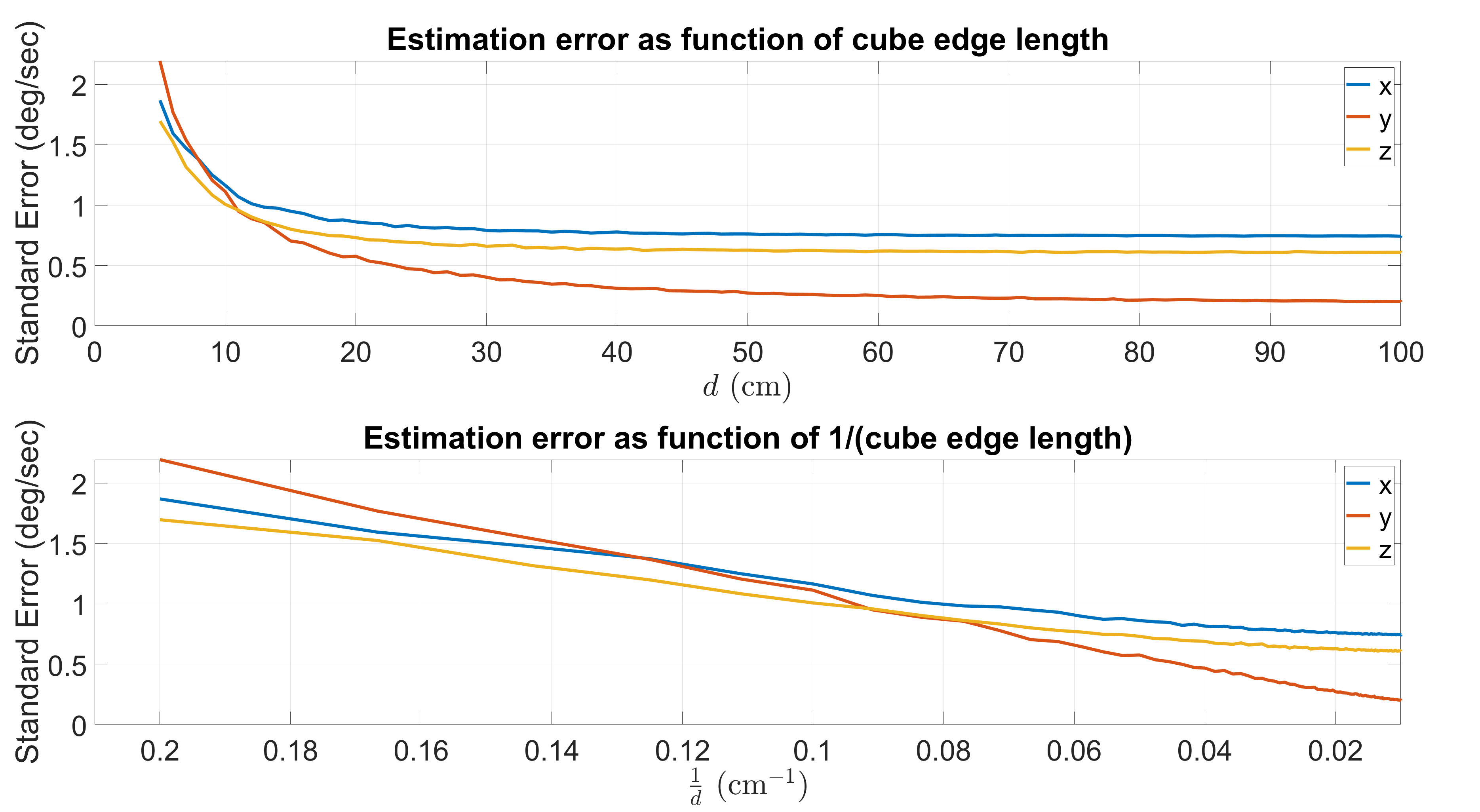}
 \end{center}
{\scriptsize \textbf{(b)}}
 \begin{center}
\includegraphics[trim={30pt 15pt 90pt 25pt},clip,width=\columnwidth]{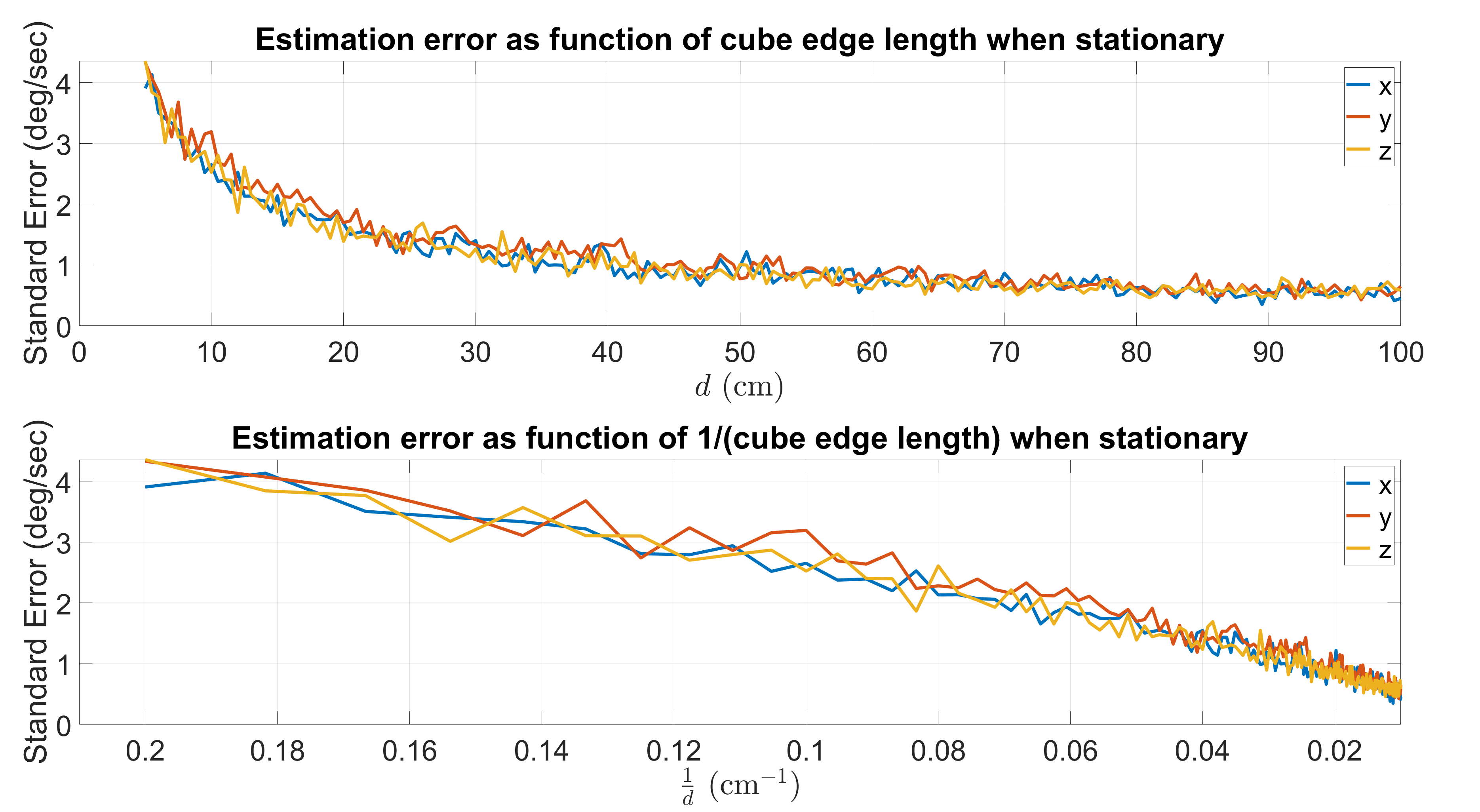}
\caption{The standard estimation error for dynamic and static simulations. The error decreases inversely with increase in the edge length $d$ of the Cu-NAA as the sensor arrangement remains unchanged.}
\label{Fig:VariableEdge}
\end{center}
\end{figure}

The effect of sensor placement on the estimator is analyzed by observing the estimate error for $100\sec$ as the cube edge $d$ is varied from $d=5cm$ to $d=100cm$ while maintaining the same accelerometer arrangement. As observed earlier, the propagation of the sensor noise depends on the relative displacement matrix $S_d$. For this Cu-NAA arrangement, analysis indicates inverse relationship between the noise and the product of singular values, edge length $d$ and is also observed in the simulation results, \Fig \ref{Fig:VariableEdge}. The standard estimation error for dynamic rotation and stationary cases decreases inversely with increase in $d$, equivalently, a linear relationship between standard error and $1/d$ is visible. 
\pagebreak
\subsection{Mechatronics of Cu-NAA}
The experimental validation is done on a Cu-NAA fabricated out of acrylic sheets. Four 6DoF MPU-6050 IMUs are housed on the inner side of the symmetric acrylic cube at (units $cm$)
\begin{align*}
    \bm{r}_1 = \begin{bmatrix}
    7.50\\-1.00\\7.61
    \end{bmatrix}, 
    \bm{r}_2 = \begin{bmatrix}
    0\\0\\0
    \end{bmatrix},     
    \bm{r}_3 = \begin{bmatrix}
    7.60\\7.30\\0.96
    \end{bmatrix},     
    \bm{r}_4 = \begin{bmatrix}
    0.15\\6.30\\8.06
    \end{bmatrix}
\end{align*}
The IMU data is retrieved using I2C communication protocol. Due to the limitations of available address names of the MPU (0x68 and 0x69), an I2C multiplexer (MUX) was used to connect all four sensors on a single I2C bus on the microcontroller. Two sensors were assigned per bus to limit the number of I2C bus switching within the MUX, thereby minimizing the lag time between the two buses and improving communication reliability. The Arduino DUE microcontroller serves as the I2C master for the entire system performing burst reads on all four accelerometers via the I2C bus, and writes these recorded values to the serial port. The serial port values are read and saved into a file using a Python script. The raw digital data is converted to accelerometer values ($m/\sec^2$) using the sensitivity and offset obtained from calibration procedure. Thereafter, the estimate is obtained using the proposed discretized EKF, \Eqn \ref{Eqn:DiscreteEKFProcess} and \ref{Eqn:DiscreteEKFMeasurement}, with the average sampling rate of $8ms$.
\begin{figure}
\begin{minipage}{0.49\columnwidth}
{\scriptsize \textbf{(a)}}
\includegraphics[trim={10pt 10pt 10pt 10pt},clip,width=\columnwidth]{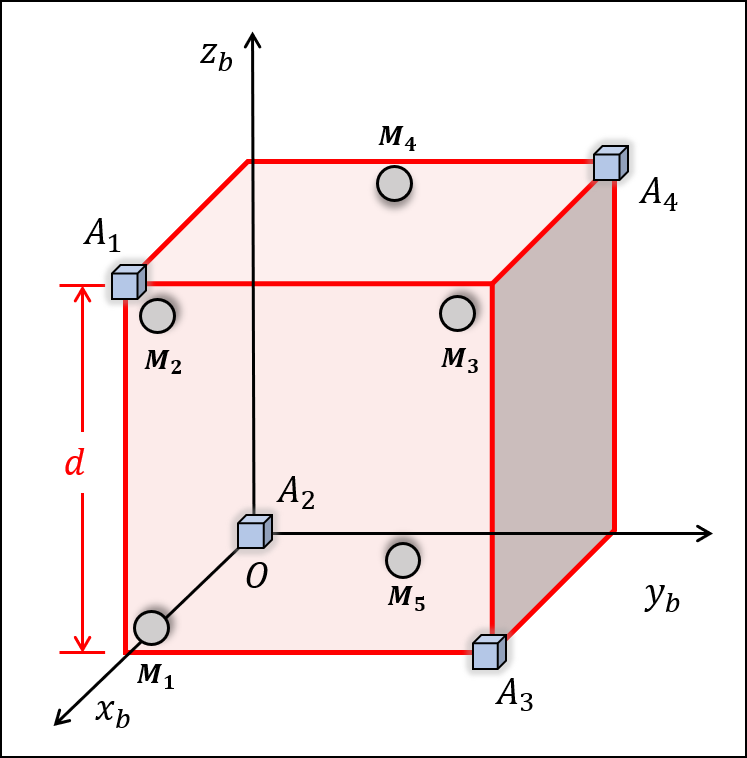}
\end{minipage}
\hfill
\begin{minipage}{0.5\columnwidth}
{\scriptsize \textbf{(b)}}
\includegraphics[trim={10pt 0pt 10pt 10pt},clip,width=\columnwidth]{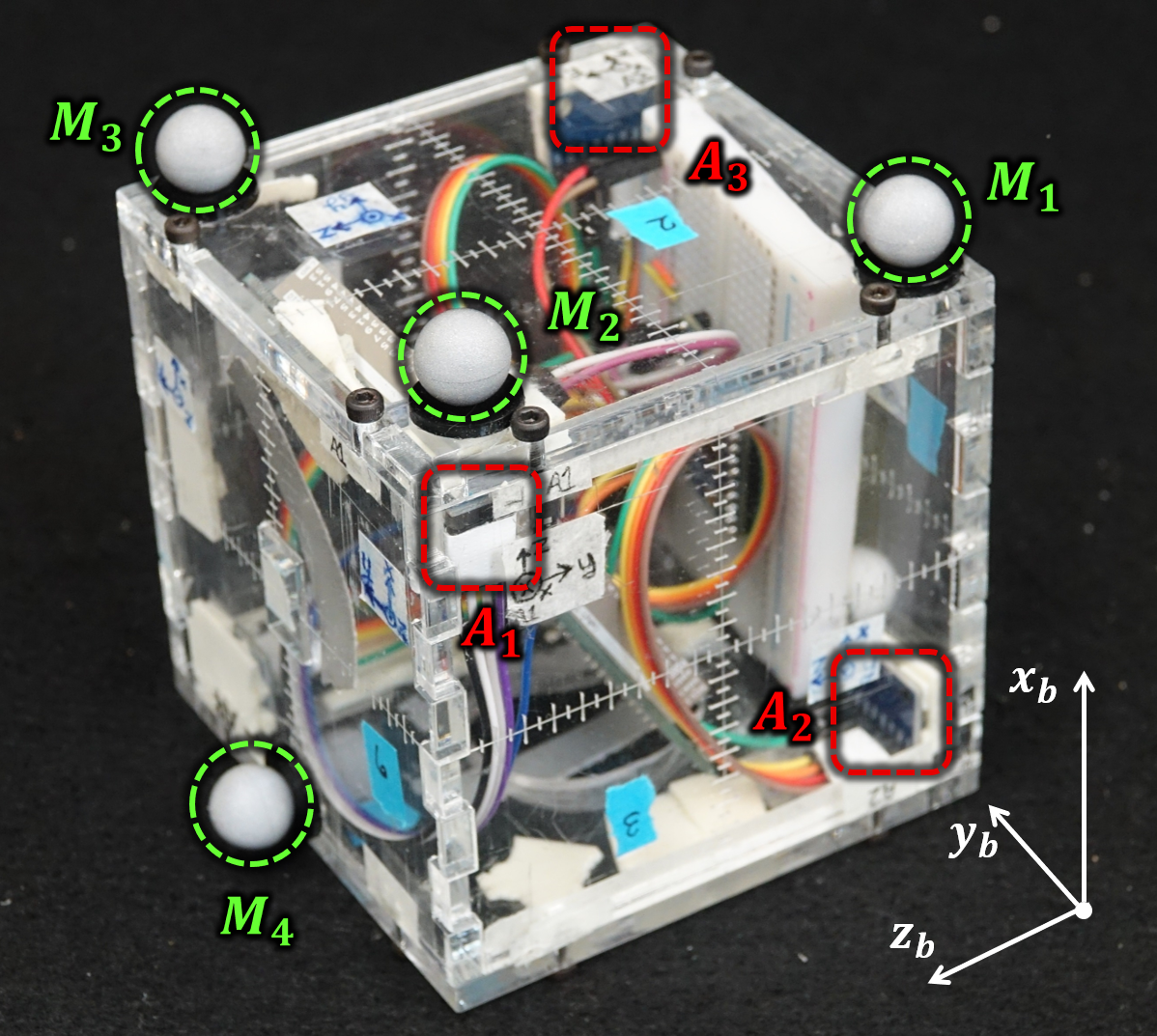}
\end{minipage}
\caption{(a) The rendering the experimental Cu-NAA with edge $d=10cm$ comprising of four accelerometers $A_i$ and optical markers $M_i,\forall i=1,\cdots,4$. (b) They are indicated by labeled red squares and blue circles respectively.}
\label{Fig:CuNAA}
\end{figure}

\begin{figure}[h]
\begin{center}
\includegraphics[trim={75pt 10pt 75pt 33pt},clip,width=\columnwidth]{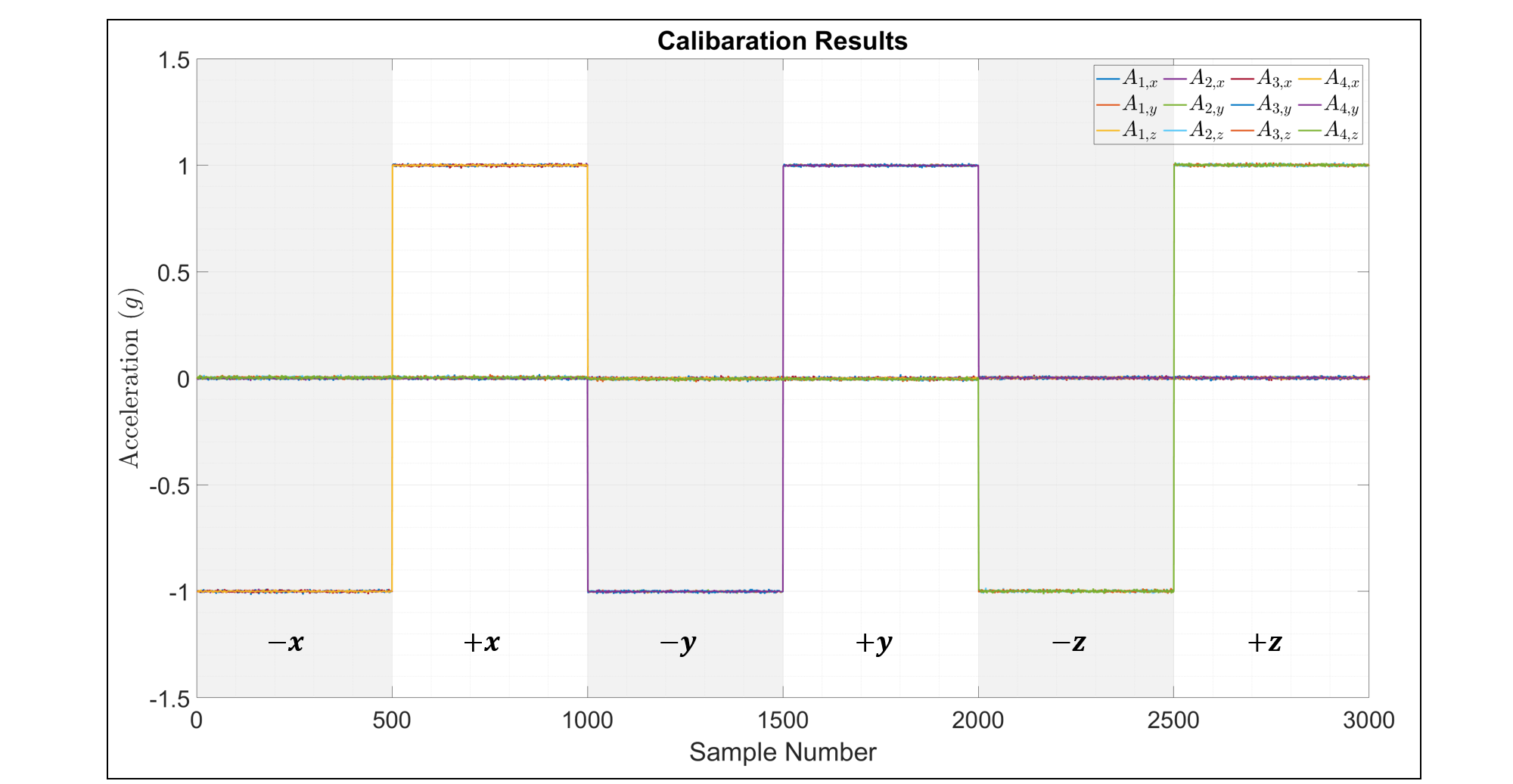}
\caption{Calibrated MEMS accelerometers where the relationship between sensor signal and acceleration is assumed to be linear. The alternately colored regions highlight different orientations where the gravity is assumed to be parallel to the respective axis of the body coordinate system. For example, the first 500 samples are correspond to the orientation when $-x_b$ is aligned with gravity.} 
\label{Fig:CalibrationPlots}
\end{center}
\end{figure}

Calibration of each individual accelerometer is performed by assuming a linear relationship between acceleration of $i$th sensor in the body coordinate system $\widehat{\bm{a}}_i$ and the sensor signal $\bm{v}_i \in \mathbb{R}^{3\times 1}$. 
\begin{align*}
    \widehat{\bm{a}}_i &= S_i\bm{v}_i +\bm{o}_i
\end{align*}
The sensitivity $S_i\in\mathbb{R}^{3\times 3}$ and offset $\bm{o}_i\in\mathbb{R}^{3\times 1}$ are obtained using linear least squares on 6 known alignments of the Cu-NAA $\pm x, \pm y, \pm z$ axes along the gravity as discussed in Appendix \ref{App:SensorCalib}. The sensors are calibrated with 500 samples in each orientation with $g=9.81 \mathrm{m}/\sec^2$, Fig. \ref{Fig:CalibrationPlots}.
\newcommand{\marker}[2]{\bm{m}_{#1}^{#2}}
\newcommand{\freev}[2]{\bm{v}_{#1}^{#2}}
\newcommand{\crossv}[2]{\bm{w}_{#1}^{#2}}
\newcommand{\thetaQUEST}{\theta_{quest}}
\subsection{Angular velocity from optical markers}
The ground truth data is obtained from the \Vicon~system with eight cameras. Five markers are placed on the experimental cube, \ref{Fig:CuNAA}b. The attitude and covariance matrices are obtained using Quarternion Estimation (QUEST) algorithm \cite{shuster_three-axis_1981}. Let $q_{\theta}(t), P_{\theta}(t)$ denote the optimal attitude quarternion and the covariance matrix obtained using QUEST. The angular velocity and the covariance matrix is evaluated as discussed in \Appen \ref{App:VICONAlgo} 
\begin{align*}
q_{\omega}(t) &= 2{q^{*}_\theta}(t)\bm{\otimes}\left(\frac{{q_\theta}(t) - {q_\theta}(t-1)}{T}\right)\\
P_{\omega}(t) &= \frac{P_{\theta}(t)+P_{\theta}(t-1)}{T^2}
\end{align*}
where ${q},{q^{*}},\bm{\times}$ denote the quarternion, its conjugate and multiplication operation. The angular velocity $\bm{\omega}(t)$ is the vector portion of $q_{\omega}(t)$.
\subsection{Experimental Results}
\begin{figure}[h]
\begin{center}
\includegraphics[trim={150pt 25pt 250pt 150pt},clip,width=\columnwidth]{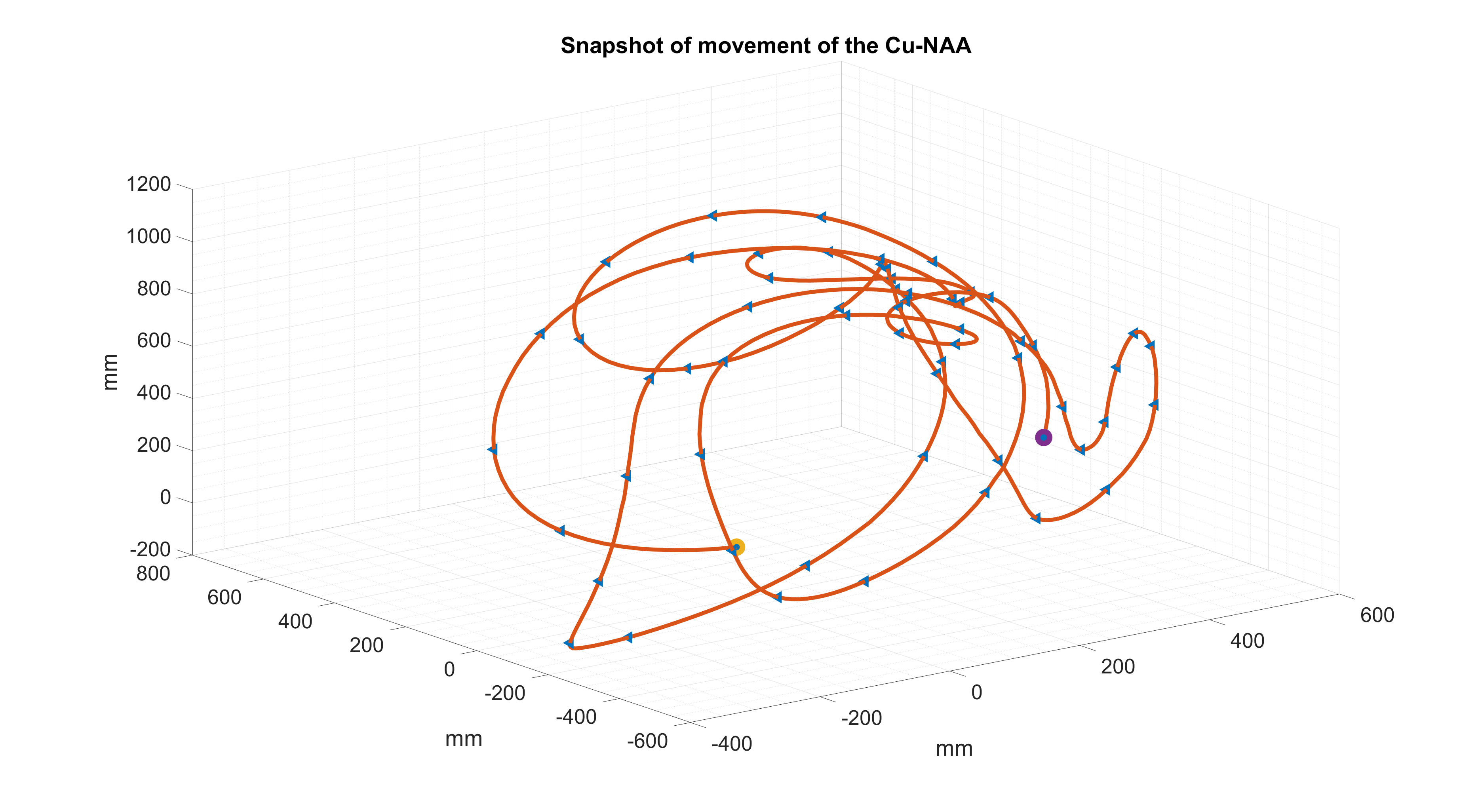}
\caption{A snapshot of movement of Cu-NAA obtained via optical tracking illustrating simultaneous rotation and translation of the sensor during the experiment.} 
\label{Fig:CuNAAMovement}
\end{center}
\end{figure}

\begin{figure*}[h]
{\footnotesize\textbf{(a)}}\vspace{-20pt}
\begin{center}
\includegraphics[trim={25pt 25pt 25pt 50pt},clip,width=1.5\columnwidth]{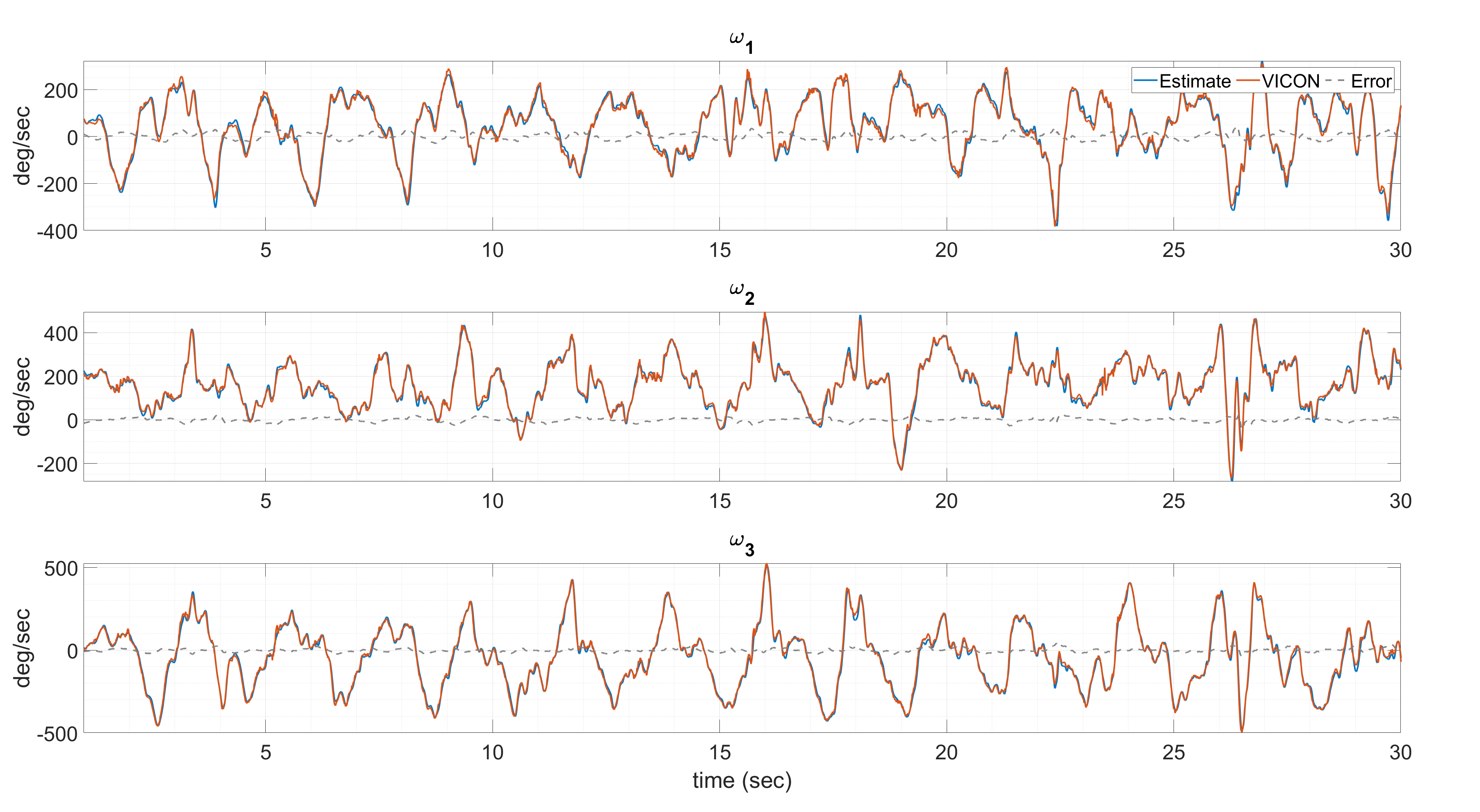}
\end{center}
{\footnotesize\textbf{(b)}}\vspace{-20pt}
\begin{center}
\includegraphics[trim={25pt 25pt 25pt 50pt},clip,width=1.5\columnwidth]{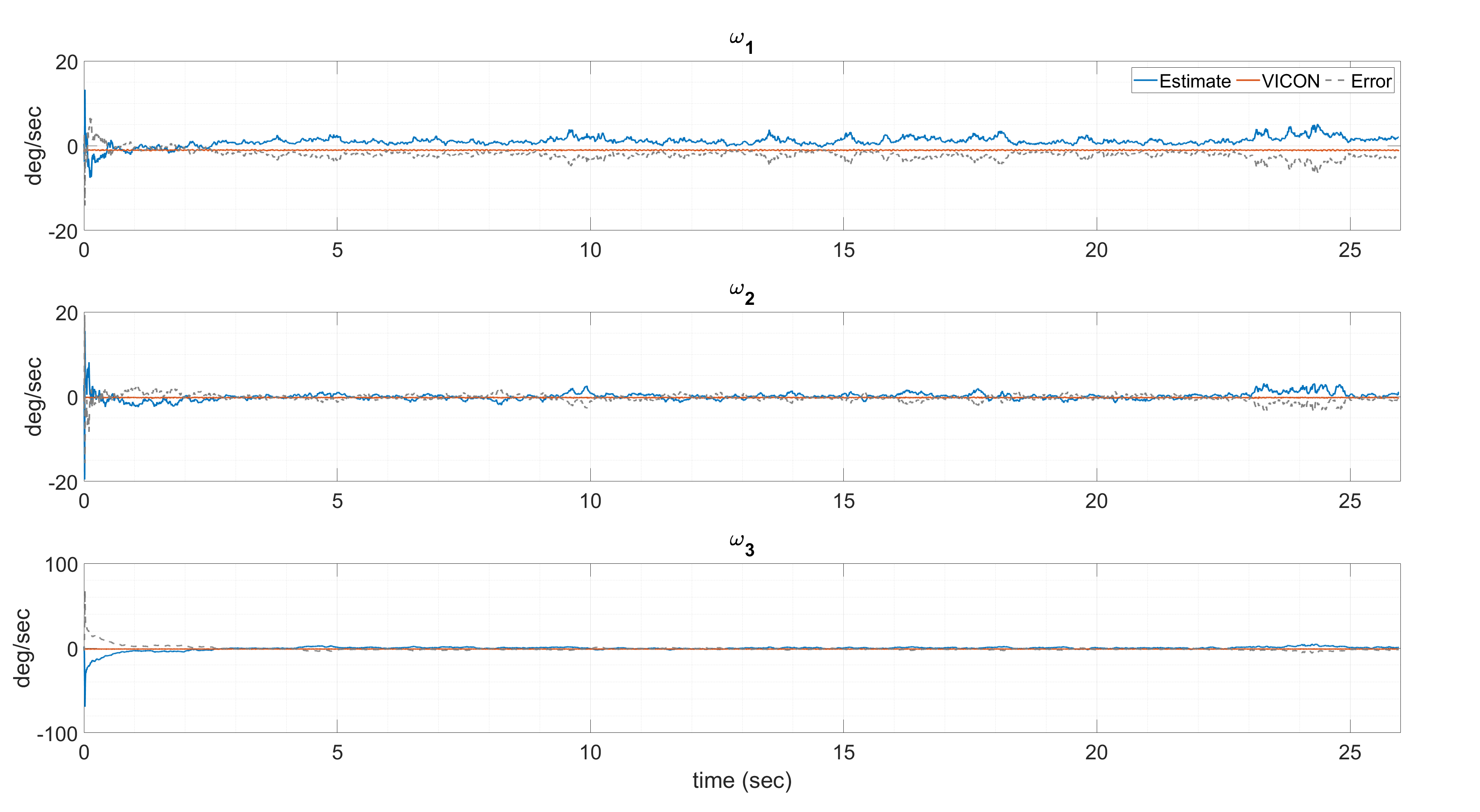}
\end{center}
\caption{Estimated (EKF) and true (optical markers using \Vicon) angular velocity in the three directions where the error is indicated in the dotted gray line. (a) A dynamic experiment yields a mean error of $0.92,-0.63,-0.96 \deg/\sec$ in the three directions. (b) Static case has mean of $-2.30,-0.39,-1.46 \deg/\sec$ with standard error of $0.80, 0.75,0.95 \deg/\sec$.} 
\label{Fig:ExperimentOmega}
\end{figure*}

During the experiments, the Cu-NAA is not fixed on an apparatus and experiences both linear and angular acceleration. An example snapshot of the Cu-NAA motion over 5 seconds is shown in Fig. \ref{Fig:CuNAAMovement}.
Extensive experiments were conducted and the results were encouraging. For example dynamic translation-rotation, \Fig \ref{Fig:ExperimentOmega}a, the EKF estimate has mean error of $0.92,-0.63,-0.96 \deg/\sec$ in the body $\{x_b,y_b,z_b\}$ directions with standard error of $13.54, 9.03, 11.31\deg/\sec$. %
The \Vicon~measurement standard error is assumed to be $2 mm$ in such dynamic environments \cite{merriaux_study_2017}. The estimated and \Vicon~angular velocity distributions may be compared using entropy metric of Kullback-Leibler divergence. However, given the anticipated systematic errors in the experiment, such comparison are not made in the current paper. While, the static case, \Fig \ref{Fig:ExperimentOmega}b, has the mean error of $-2.30, -0.39, -1.46 \deg/\sec$ with standard error of $0.80, 0.75, 0.95 \deg/\sec$. In comparison, the EKF estimator with correlated process and measurement noise ($L=\bm{0}_{3\times 6}$ in \Eqn {\ref{Eqn:Sysprocess}}) has mean and standard error of (i) dynamic case: $-1.17, 2.93, 4.28 \deg/\sec$,  $14.09,9.17, 11.81 \deg/\sec$, and (ii) static case:  $-1.92, $ $1.31, 2.01\deg/\sec$, $1.56, 1.43, 2.70 \deg/\sec$ respectively in the body $\{x_b,y_b,z_b\}$. The experimental results indicate that the proposed EKF with uncorrelated noise performs better than the one with correlated process-measurement noise.
The sources of error are anticipated to be systematic in nature - placement of sensors ($S_d$ matrix) and calibration of accelerometers (sensitivity and bias).

%% file: 05-Conclusion.tex
\section{Conclusion}\label{Sec:Conclusion}
We estimate angular velocity using a Non-coplanar Accelerometer Array (NAA) where four or more triaxial accelerometers are arranged in non-specific orientation with the geometric constraint of non-coplanarity. Mathematically, this sensor arrangement is reflected in the relative displacement matrix. The singular values of this matrix - more specifically, their product and condition number, determine the propagation of the sensor noise and the noise covariance matrix. %
A continuous-discrete dynamical system model is derived where the accelerometer readings are used simultaneously as process input and the measurement with uncorrelated process and measurement noises. The angular velocity is estimated using an Extended Kalman Filter (EKF) that discretizes and linearizes the nonlinear system model. %
The simulations and experiments are performed on a Cube-NAA (Cu-NAA) comprising of four accelerometers placed at different vertices of a cube, and five reflective optical markers. For static and dynamic cases, the simulation results are encouraging and show an asymptotic decrease in the standard estimation error with distance as the sensor arrangement is kept constant and the edge length of the cube is increased. Experimentally, the ground truth angular velocity is determined using the reflective markers that are visually tracked using \Vicon. This unique experimental setup is capable of providing six degrees of freedom - three degrees each for rotational and spacial translation (linear acceleration of the Cu-NAA in space) freedom of movement. The experimental results for Cu-NAA with edge length of $10cm$ are extremely encouraging and the errors are anticipated to be systematic in nature - placement of the sensors and sensor calibration. Overall, the simulation and experimental results indicate better performance of the proposed EKF with uncorrelated process and measurement noises, as compared to the one without. In future, the systematic errors can be solved by incorporating the constant parameters into the system dynamics.

%% file: 06-Appendices.tex
\appendices 
\section{Accelerometer calibration using linear least squares}
\label{App:SensorCalib}
For a given sensor, the linear relationship between the acceleration $\bm{a}$ and sensor signal $\bm{v}$ is defined using sensitivity $S$ and offset $\bm{o}$
\begin{align*}
    \widehat{\bm{a}} = S\bm{v} + \bm{o}
\end{align*}
for $S\in\mathbb{R}^{3\times 3}$ and $\widehat{\bm{a}},\bm{v},\bm{o}\in \mathbb{R}^{3\times 1}$. This is re-written as
\begin{align*}
    \widehat{\bm{a}}&=V(\bm{v})\bm{y}\\
    V(\bm{v}) &=\begin{bmatrix}
    \bm{v}^T & 0 &0 &1 &0 &0\\
    0 & \bm{v}^T &0 & 0&1 &0\\
    0 & 0 & \bm{v}^T& 0 &0 &1
    \end{bmatrix}\\
    \bm{y} &= \left[S_{11},S_{12},S_{13},%
    S_{21},S_{22},S_{23},S_{31},S_{32},S_{33},%
    o_1,o_2,o_3\right]^T
\end{align*}
Accelerometer readings are taken for 6 known orientations $\pm x,\pm y,\pm z$ w.r.t body (Cu-NAA) coordinate system. It can be observed that only four linearly independent orientations are required for obtaining the unknowns. Consequently, the calibration constants, $\bm{y}$, are calculated using linear least squares solution
\newcommand{\sensor}{\widehat{\bm{a}}}
\newcommand{\VMat}[1]{V(\bm{v}_{#1})}
\begin{align*}
    \bm{y}&=\mathcal{V}^{+}\mathcal{A}\\
    \mathcal{A} &= [\sensor_1^T,\sensor_2^T,\cdots,\sensor_N^T]^T\\
    \mathcal{V} &= \left[\VMat{1}^T, \VMat{2}^T,\cdots,\VMat{N}^T\right]^T
\end{align*}
where $\mathcal{A}\in\mathbb{R}^{3N\times 1}, \mathcal{V}\in \mathbb{R}^{3N\times 12}$ and the superscript ${+}$ denotes the pseudoinverse of the matrix.
\section{Obtaining body angular velocity using visual tracking of markers on rigid body} \label{App:VICONAlgo}
\begin{figure}[h]
\begin{center}
\includegraphics[trim={5pt 5pt 10pt 10pt},clip,width=0.7 \columnwidth]{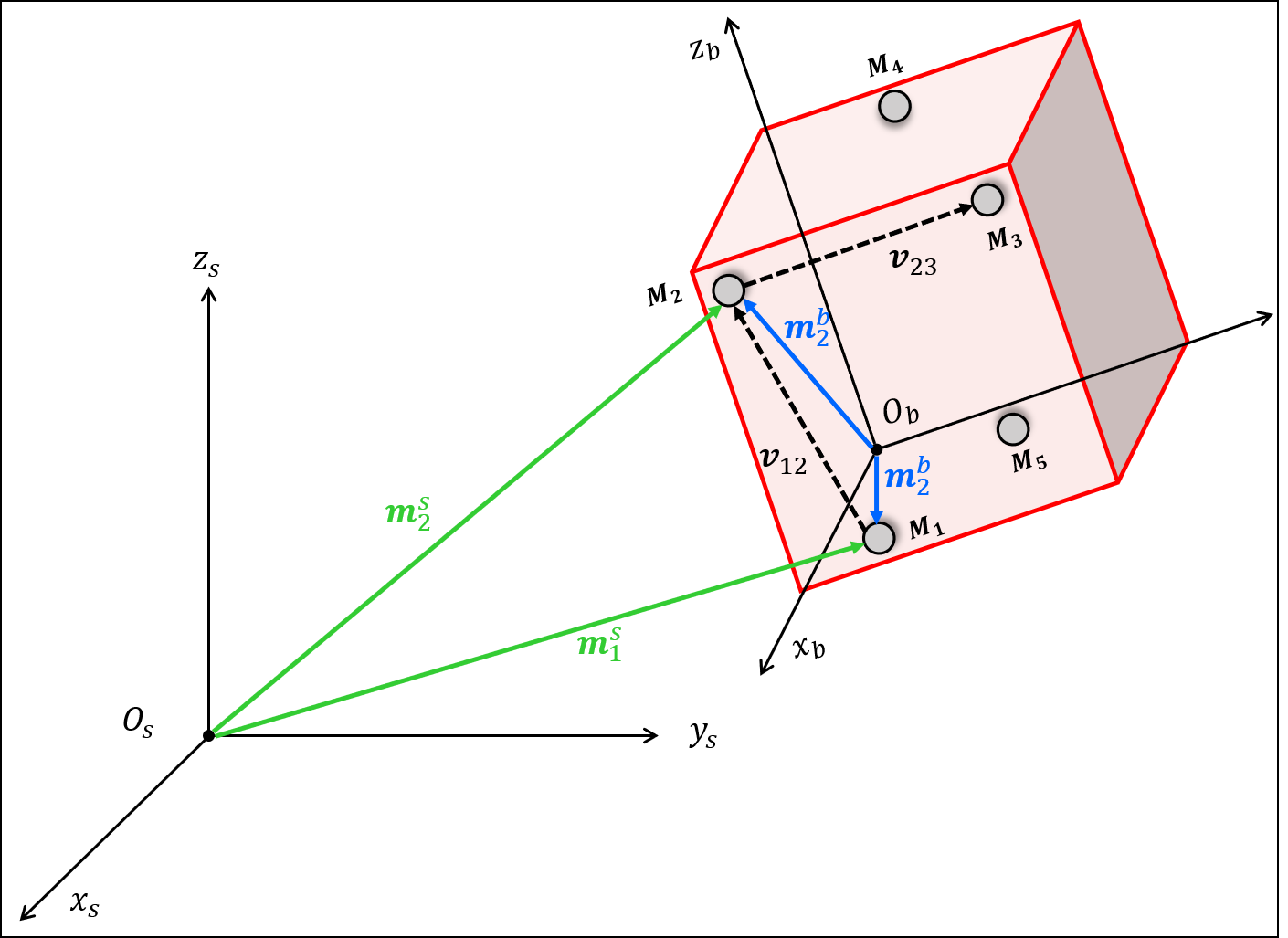}
\caption{The \Vicon~markers $M_i$ and the vector representations in body and inertial coordinate systems.}
\label{Fig:VICONTracking}
\end{center}
\end{figure}

Let the position of $N$ markers on a rigid body be known in the body $\{b\}$ and inertial $\{s\}$ coordinate systems where $\marker{i}{a}$ denotes the position of $i$th marker in $a$ coordinate system, \Fig \ref{Fig:VICONTracking}. Additionally, let $R$ be the rotation matrix between $\{s\}$ and $\{b\}$. The QUEST algorithm\cite{shuster_three-axis_1981} is a solution to the Wahba's problem that finds the attitude by minimizing the cost function
\begin{align*}
J(A) &= \frac{1}{2}\sum_{i=1}^{N}w_i\left|\left|\marker{i}{b}-R\marker{i}{s}\right|\right|^2
\end{align*}
by solving the quarternion eigenvalue equation. The weights are chosen as $w_i=1/\sigma_i^2$ where $\sigma_i^2$ is the variance of the measurement vectors. Succinctly, the angular velocity and convariance matrices are obtained in the following sequential manner.
\begin{enumerate}
\item Define attitude profile matrix $B$, and quantities $S,Z,\sigma,\kappa, \Delta$
\begin{align*}
B &= \sum_{i=1}^{N}\marker{i}{b}{\marker{i}{s}}^T, \quad S=B+B^T\\
\sigma&=\mathrm{tr}(B), \kappa = \mathrm{tr}\left(\mathrm{adj}(S)\right), \Delta = \mathrm{det}(S)\\
Z &= \left[B_{23}-B_{32}, B_{31}-B_{13}, B_{12}-B_{21}\right]^T
\end{align*}
\item Obtain maximum eigenvalue $\lambda_{max}$ by solving the characteristic equation
\begin{align*}
&\lambda^4  + (a+b)\lambda^2 - c\lambda +(ab+c\sigma-d) = 0\\
&\mathrm{s.t.}\quad a = \sigma^2 -\kappa, \quad b = \sigma^2 +Z^TZ \\
&\qquad c = \Delta+ Z^TSZ, \quad d = Z^TS^2Z
\end{align*}
\item Construct optimal attitude quarternion and covariance matrix
\begin{align*}
q_{opt,\theta} &= \frac{1}{\sqrt{\gamma^2 +|| \bm{x}||^2}}\begin{bmatrix}
\gamma\\\bm{x}
\end{bmatrix}\\
& \mathrm{s.t.} \quad \bm{x} = (\alpha I +\beta S + S^2)Z, \\
&\alpha = \lambda_{max}^2 -\sigma^2+\kappa, \beta = \lambda_{max}-\sigma\\
& \gamma = (\lambda_{max}+\sigma)\alpha - \Delta\\
P_{\theta} &= \left[\sum_{i=1}^N\frac{1}{\sigma_i^2}\left(I - \marker{i}{b}{\marker{i}{b}}^T\right)\right]^{-1}
\end{align*}
where $\sigma_i^2$ is the variance of the $i$th measurement.
\item Calculate angular velocity and the covariance matrix
\begin{align*}
q_{\omega}(t) &= 2{q^{*}_\theta}(t)\bm{\otimes}\left(\frac{{q_\theta}(t) - {q_\theta}(t-1)}{T}\right)\\
P_{\omega}(t) &= \frac{P_{\theta}(t)+P_{\theta}(t-1)}{T^2}
\end{align*}
where ${q},{q^{*}},\bm{\times}$ denote the quarternion, its conjugate and multiplication operation. The angular velocity $\bm{\omega}(t)$ is the vector portion of $q_{\omega}(t)$.
\end{enumerate}

